\documentclass[11pt,a4paper]{article}
\usepackage[hyperref]{acl2020}
\usepackage{times}
\usepackage{latexsym}
\usepackage{graphicx}
\usepackage{booktabs}
\usepackage{xcolor}
\usepackage{tcolorbox}
\usepackage{url}  

\usepackage{array}
\newcolumntype{L}[1]{>{\raggedright\let\newline\\\arraybackslash\hspace{0pt}}m{#1}}
\newcolumntype{C}[1]{>{\centering\let\newline\\\arraybackslash\hspace{0pt}}m{#1}}
\newcolumntype{R}[1]{>{\raggedleft\let\newline\\\arraybackslash\hspace{0pt}}m{#1}}
\usepackage{enumitem}
\definecolor{OliveGreen}{rgb}{0,0.6,0}
\newcommand\Mark[1]{\textsuperscript#1}

\usepackage{microtype}

\aclfinalcopy 

\title{MultiWOZ 2.2 : A Dialogue Dataset with Additional Annotation Corrections and State Tracking Baselines}

\author{Xiaoxue Zang\Mark{1}, Abhinav Rastogi\Mark{1}, Srinivas Sunkara\Mark{1}, Raghav Gupta\Mark{1},\\
 \bf{Jianguo Zhang}\Mark{2}, \bf{Jindong Chen}\Mark{1} \\
 \Mark{{1}}Google Research, \Mark{2}University of Illinois at Chicago\\
  \Mark{1}\texttt{\{xiaoxuez, abhirast, srinivasksun, raghavgupta\}@google.com},\\
  \Mark{2}\texttt{jzhan51@uic.edu}, \Mark{1}\texttt{jdchen@google.com}\\
 }

\date{}

\begin{document}
\maketitle
\begin{abstract}
MultiWOZ \cite{budzianowski2018multiwoz} is a well-known task-oriented dialogue dataset containing over 10,000 annotated dialogues spanning 8 domains. It is extensively used as a benchmark for dialogue state tracking. However, recent works have reported presence of substantial noise in the dialogue state annotations. MultiWOZ 2.1 \cite{eric2019multiwoz} identified and fixed many of these erroneous annotations and user utterances, resulting in an improved version of this dataset. This work introduces MultiWOZ 2.2, which is a yet another improved version of this dataset. Firstly, we identify and fix dialogue state annotation errors across 17.3\% of the utterances on top of MultiWOZ 2.1. Secondly, we redefine the ontology by disallowing vocabularies of slots with a large number of possible values (e.g., restaurant name, time of booking). In addition, we introduce slot span annotations for these slots to standardize them across recent models, which previously used custom string matching heuristics to generate them. We also benchmark a few state of the art dialogue state tracking models on the corrected dataset to facilitate comparison for future work. In the end, we discuss best practices for dialogue data collection that can help avoid annotation errors. 
\end{abstract}

\section{Introduction}
Task-oriented dialogue systems have become very popular in the recent years. Such systems assist the users in accomplishing different tasks by helping them interact with APIs using natural language. Dialogue systems consist of multiple modules which work together to facilitate such interactions. Most architectures have a natural language understanding and dialogue state tracking module to generate a structured representation of user's preferences from the dialogue history. This structured representation is used to make API calls and as a signal for other modules. Then, the dialogue policy module determines the next actions to be taken by the dialogue system. This is followed by the natural language generation module, which converts the generated actions to a natural language utterance, which is surfaced to the user.

Recently, data-driven techniques have achieved state-of-the-art performance for the different dialogue systems modules~\cite{Wen_2017, Ren_2018, zhang2019find,chao2019bert}. However, collecting high quality annotated dialogue datasets remains a challenge for the researchers because of the extensive annotation required for training the modules mentioned above. Many public datasets like DSTC2 \cite{henderson2014second}, WOZ~\cite{wen2017network}, SimulatedDialogue~\cite{shah2018building}, MultiWOZ~\cite{budzianowski2018multiwoz}, TaskMaster~\cite{byrne2019taskmaster}, SGD~\cite{rastogi2019scalable}, etc. have been very useful to facilitate research in this area. Among these datasets, MultiWOZ is the most widely used benchmark for dialogue state tracking. It contains over 10,000 dialogues spanning 8 domains, namely - Restaurant, Hotel, Attraction, Taxi, Train, Hospital, Bus, and Police. 

Since its inception, the MultiWOZ dataset has undergone a few updates. \citet{lee2019convlab} introduced user dialogue actions providing a structured semantic representation for user utterances. \citet{eric2019multiwoz} fixed 32\% of dialogue state annotations across 40\% of the turns and introduced slot descriptions, culminating in MultiWOZ 2.1, a new version of the dataset. Despite the large scale of corrections introduced in MultiWOZ 2.1, there are still many unaddressed annotation errors ~\cite{zhang2019find}. Furthermore, several approaches to dialogue state tracking use span annotations identifying the locations in the user and system utterances where slot values have been mentioned, to make the system efficient and generalizable to new slot values~\cite{rastogi2017scalable,WuTradeDST2019,zhang2019find,rastogi2019scalable, xu-hu-2018-end,zhou2019multidomain, Gao_2019}. Because of the absence of these span annotations in MultiWOZ, these approaches resort to generating them using custom string matching heuristics, making their comparison difficult. 

To address these limitations, we introduce MultiWOZ 2.2\footnote{The dataset is available at \url{https://github.com/budzianowski/multiwoz}.}, an updated version of the MultiWOZ dataset. Our contributions are threefold. 
\begin{enumerate}[leftmargin=*]
    \item We identify the annotation errors, inconsistencies, and ontology issues in MultiWOZ 2.1, and publish its improved version.
    \item We add slot span annotations for user and system utterances to standardize them across future models. We also annotate the active user intents and requested slots for each user utterance.
    \item We benchmark a few state-of-the-art dialogue state tracking models on the corrected dataset to facilitate comparison for future work.
\end{enumerate}

The paper is organized as follows. First we describe the different types of annotation errors and inconsistencies we observed in MultiWOZ 2.1 (Section \ref{sec:issues}). Then, we outline the redefinition of ontology (Section \ref{sec:ontology}), followed by the description of correction procedure (Section \ref{sec:correction}) and new annotations we introduce (Section~\ref{sec:additional}). Finally, in Section \ref{sec:benchmark}, we present the performance of a few recent dialogue state tracking models on MultiWOZ 2.2.

\begin{figure*}[t]
\centering
\begin{minipage}{15.6cm}\vspace{0mm}    \centering
\begin{tcolorbox}[colback=yellow!5!white]
    \centering
    \small
    \begin{tabular}{ L {61mm}  C {35mm}  C {35mm}}
     \textbf{Example Dialogue Segment}& \textbf{MultiWOZ 2.1} & \textbf{MultiWOZ 2.2} \\
    \midrule
    \textbf{1. Early Markup} & & \\
        \textcolor{blue}{User:} Help me find a moderate priced british food place please. & r-food=british, r-pricerange=moderate, \textcolor{red}{r-name=one seven} & r-food=british, r-pricerange=moderate  \\
    \textcolor{blue}{Sys:} restaurant one seven is a nice place. Do you want to book? & & \\ \midrule
    
    \textbf{2. Annotation from Database} & & \\[3pt]
    \textcolor{blue}{User:} Can you give me the address to the hospital in Cambridge? & \textcolor{red}{hospital-department=acute medical assessment unit} & -no update- \\
    \textcolor{blue}{Sys:} The address is Hills Rd, Cambridge Postcode: CB20QQ & & \\ \midrule
    
    \textbf{3. Typo} & & \\
    \textcolor{blue}{Sys:} Okay, I can help with that. What day and time would you like to dine and how many people should I have the reservation for? & r-bookday=thursday, \textcolor{red}{r-booktime=15:00}, hotel-area=west & r-bookday=thursday, \textcolor{OliveGreen}{r-booktime=5:00}, hotel-area=west\\
    \textcolor{blue}{User:} On Thursday at 5:00. I also need a hotel in the same area. No need to have free parking. & & \\ \midrule
    
    \textbf{4. Implicit Time Processing} & & \\
    \textcolor{blue}{User:} Can I get the postcode for that? I also need to book a taxi to the Golden Wok. & r-name=Golden Wok, r-bookday=friday, r-booktime=11:00, \textcolor{red}{taxi-leaveAt=friday}, taxi-destination=Golden Wok& r-name=Golden Wok, r-bookday=friday, r-booktime=11:00, taxi-destination=Golden Wok \\
    \textcolor{blue}{Sys:} The postcode is cb21tt. Are you looking for a taxi from Old Schools to the Golden Wok? & & \\
    \textcolor{blue}{User:} Yes I do. I'd like to make sure I arrive at the restaurant by the booked time. Can you check? & r-name=Golden Wok, r-bookday=friday, r-booktime=11:00, \textcolor{red}{taxi-leaveAt=friday}, \textcolor{red}{taxi-arriveby=10:45} & r-name=Golden Wok, r-bookday=friday, r-booktime=11:00, \textcolor{OliveGreen}{taxi-arriveby=11:00}\\
 \vspace{-5mm}
    \end{tabular}
\end{tcolorbox}
\vspace{-2mm}
\caption{Examples of hallucinated values in MultiWOZ 2.1 and the corrections in MultiWOZ 2.2. Please note that we omit state annotations unrelated to the extracted utterances. ``r" used in the slot name in the right two columns is an abbreviation of restaurant.}
    \label{fig:ann_error}
\end{minipage}
\end{figure*}

\section{Annotation Errors}\label{sec:issues}
The MultiWOZ dataset was collected using a Wizard-of-Oz setup~\cite{kellywoz1984}. In this setup, two crowd-workers are paired together, one acting as a user and the other as the dialogue agent. Each dialogue is driven by a unique set of instructions specifying the user goal, which are shared with the crowd-worker playing the role of the user. After every user turn, the crowd-worker playing the role of the dialogue agent (wizard) annotates the updated dialogue state. After updating the state, the tool shows the set of entities matching the dialogue state to the wizard, who then uses it to generate a response which is sent to the user. Remaining annotations such as the system actions are collected using a second annotation task.

The Wizard-of-Oz setup is widely considered to produce natural conversations, as there is no turn level intervention guiding the flow of the dialogue. However, because of its heavy reliance on humans for generating the correct annotations, the procedure is prone to noisy annotations. We identified two major classes of errors outlined below, which were not corrected in MultiWOZ 2.1.

\subsection{Hallucinated Values}
\label{sec:hallucinated}
Hallucinated values are present in dialogue state without being specified in the dialogue history. We observed four different types of such errors, which are shown in Figure~\ref{fig:ann_error} and described below. 

\begin{enumerate}[style=nextline,leftmargin=*]
    \item \textbf{Early Markups:} These values have been mentioned by the agent in a future utterance. Since the user has not accepted them yet, they should be excluded from the dialogue state.
    \item \textbf{Annotations from Database:}
    These values are not mentioned in the dialogue at all, even in the future utterances. They appear to be incorrectly added by the wizard based on results of the database call.
    \item \textbf{Typos:} These values cannot be found in the dialogue history because of a typographical error. These errors occur since slot values are entered as free-form text in the annotation interface.
    \item \textbf{Implicit Time Processing:} This specifically relates to slots taking time as a value. Sometimes, the value is obtained by adding or subtracting some pre-determined duration from the time specified in dialogue history (Figure~\ref{fig:ann_error}). In other cases, it is implicitly rounded off to closest quarter (Dialogue 1 in Figure~\ref{fig:inconsis_update}). This further burdens models with learning temporal arithmetic.
\end{enumerate}

We observed that the errors mentioned above are quite frequent.
In total we found that hallucinated values appear in 3128 turns across 948 dialogues in the MultiWOZ 2.1 dataset.

\subsection{Inconsistent State Updates}
\label{sec:inconsis_issues}

\begin{table*}[ht]
\centering
    \begin{tabular}[t]{|L {20mm} | L {60mm} | L {50mm}|}\hline
            \textbf{Source} & \textbf{User utterance} & \textbf{Dialogue state update}\\\hline
            Ontology & I need to arrive by 8:00. & train-arriveby=08:00\\ \hline
            Dialogue history& Sometime after 5:45 PM would be great.& train-leaveat=5:45pm\\ \hline
            None & I plan on getting lunch first, so sometime after then I'd like to leave. & train-leaveat=after lunch\\ \hline
    \end{tabular}
    \caption{Example of slot values annotated using different strategies in ``PMUL0897.json", `MUL0681.json`", and ``PMUL3200.json" in MultiWOZ 2.1.}
    \label{table:diff_anns}
\end{table*}

We also encountered annotations in MultiWOZ 2.1 that are semantically correct, but don't follow consistent annotation guidelines. Inconsistencies arise in the dialogue state because of three main reasons:
\begin{enumerate}[leftmargin=*]
    \item \textbf{Multiple Sources:} A slot value may be introduced in the dialogue state through various sources. It may either be mentioned by the user, offered by the system, carried over from another slot in the dialogue state of a different domain, or be a part of the ontology.
    \item \textbf{Value Paraphrasing:} The same slot value can be mentioned in many different ways, often within the same dialogue e.g. the value ``18:00" for the slot time may be mentioned as ``6 pm", ``1800", ``0600 pm", ``evening at 6" etc.
    \item \textbf{Inconsistent tracking strategy:} Crowd-workers have inconsistent opinions on which slot values should be tracked in the same dialogue context. For example, some workers track all slot values that the user agrees with while others only track user-specified slot values.
\end{enumerate}

Table~\ref{table:diff_anns} shows dialogue state update from three different sources for similar slots from different dialogues in MultiWOZ 2.1. In the first case, the value ``08:00" for slot \textit{train-arriveby} comes from the ontology, despite the presence of an equivalent value ``8:00" in the user utterance. On the other hand, in the second example, the slot value in the dialogue state comes from the user utterance despite the ontology listing ``17:45" as a value for the slot \textit{train-leaveat}. In the third example, the value of \textit{train-leaveat} is not derived from any of the sources mentioned above, but is generated by incorporating the semantics. The slot value can be mentioned in multiple ways, but in order to evaluate a dialogue system fairly, it's necessary to either maintain a consistent rule for deciding how the value is picked among all the mentions or consider all the mentions as the correct answer. MultiWOZ 2.1 gives one unique correct answer for each dialogue state but lacks an explicit rule on how it is determined. This inconsistency confuses the model during training and unfairly penalizes it during evaluation if it outputs a slot value which is different but equivalent to the one listed in ground truth.

\begin{figure*}[t]
\centering
\includegraphics[width=0.98\textwidth]{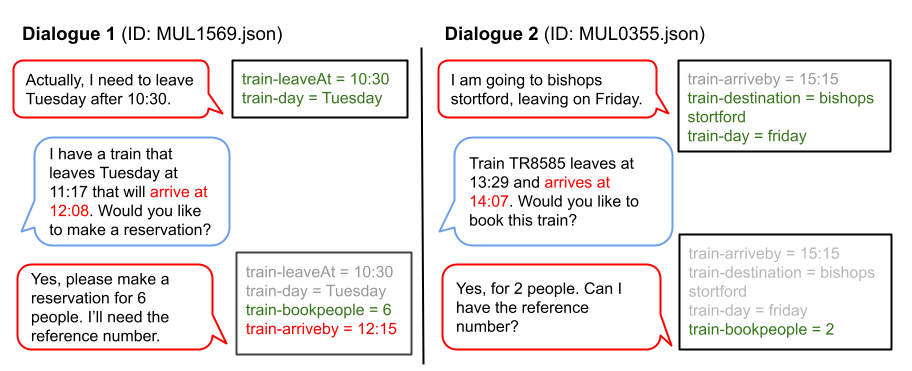}
\caption{Example of dialogues states being updated differently in similar scenarios. In both dialogues, user accepts a train offered by the system. In dialogue 1, \textit{train-arriveby} is annotated in the dialogue state after user's agreement, but not in dialogue 2. Dialogue 1 also shows implicit time processing, where the value 12:08 in the system utterance is rewritten to 12:15 in the subsequent dialogue state.}
\label{fig:inconsis_update}
\end{figure*}

Figure~\ref{fig:inconsis_update} shows another example where dialogue states are updated differently in similar scenarios. In both dialogues, the system offers an instance that fulfills the user's requirement, but the dialogue states are updated differently after user shows an intent to book the ticket. Specifically, in dialogue 1 the value for \textit{train-arriveby} provided by the system is tracked in the dialogue state while not in dialogue 2. Dialogue 1 also showcases the implicit time processing issue discussed in Section \ref{sec:hallucinated}, where the time ``12:08" has been rounded to ``12:15" in the dialogue state.

\begin{table*}[]
\centering
    \begin{tabular}[t]{|C {20mm} | L{60mm}  | L{60mm} |} \hline
        \textbf{Domain} & \textbf{Categorical slots} & \textbf{Non-categorical slots} \\ \hline
        Restaurant & pricerange, area, bookday, bookpeople & food, name, booktime \\ \hline
        Attraction & area, type & name \\ \hline
        Hotel & pricerange, parking, internet, stars, area, type, bookpeople, bookday, bookstay & name \\ \hline
        Taxi & - & destination, departure, arriveby, leaveat \\ \hline
        Train & destination, departure, day, bookpeople & arriveby, leaveat \\ \hline
        Bus & day & departure, destination, leaveat \\ \hline
        Hospital & - & department \\ \hline
        Police & - & name \\ \hline
    \end{tabular}
    \caption{Categorical and non-categorical slots defined for 8 domains in MultiWOZ 2.2.}
    \label{table:schema}
\end{table*}

\section{Ontology Issues}
\label{sec:ontology}
Although MultiWOZ 2.0 provides a predefined ontology which is claimed to enumerate all slots and the possible values for every slot, it has been reported to be incomplete. As a result, many researchers have built their own ontology to achieve a better performance~\cite{WuTradeDST2019, Goel_2019}. To fix the problem of incompleteness, MultiWOZ 2.1 rebuilt the ontology by listing all values present in dialogue states across the dataset, but it still has some unaddressed issues.

First, for some slots, multiple values sharing the same semantics are listed. Some examples are ``8pm" and ``20:00", ``a and b guesthouse" and ``a and b guest house", ``cheap$|$moderate" and ``moderate$|$cheap" for the slots \textit{restaurant-book-time}, \textit{hotel-semi-name} and \textit{hotel-semi-pricerange} respectively. We find that 51\% of the values for the slot \textit{hotel-name} are not semantically unique, and similar figures for the \textit{restaurant-name} and \textit{attraction-name} slots. Such duplicate values make evaluation hard since MultiWOZ 2.1 only assumes one correct value for each slot in the dialogue state.

Second, we observe multiple slot values in the ontology that can't be associated with any entities in the database. Values like ``free" for slot \textit{attraction-name}; ``cam", ``dif", and ``no" for slot \textit{restaurant-name} are some examples. Such values could be introduced in the ontology because of typographical errors in the utterances or annotation errors. Our investigation shows that 21.0\% of the slot values in the ontology can't be directly mapped back to the values in the database through exact string matching. We also observed a few logical expressions like ``cheap$|$moderate", ``NOT(hamilton lodge)" etc. in the ontology. We believe that these expressions, although semantically correct, add noise during training. The ontology should either omit such expressions altogether or include all possible expressions to enable generalization to cases not observed in the training data.

\section{Correction Procedure}
\label{sec:correction}

To avoid the issues described above, we advocate the definition of ontology prior to data collection. This not only serves as a guideline for annotators, but also prevents annotation inconsistencies in the dataset and corruption of the ontology from typographical and annotation errors. This section describes our definition of the new ontology, which we call \textit{schema}, followed by the corrections made to the state and action annotations. Lastly, we also show the statistics of our modifications.

\subsection{Schema Definition}
It is not realistic for the ontology to enumerate all the possible values for some slots like \textit{restaurant-name} and \textit{restaurant-booktime}, which can take a very large set of values. With addition or removal of entities in the database, the set of possible values also keeps changing continuously. \citet{rastogi2019scalable} proposed a representation of ontology, called \textit{schema}, to facilitate building a scalable dialogue system that is capable of handling such slots. A \textit{schema} divides the different slots into two categories - \textit{non-categorical} and \textit{categorical}. Slots with a large or dynamic set of possible values are called \textit{non-categorical}. Unlike ontology, the schema doesn't provide a pre-defined list of values for such slots. Their value is extracted from the dialogue history instead.

On the other hand, slots like \textit{hotel-pricerange} or \textit{hotel-type}, which naturally take a small finite set of values are called \textit{categorical}. Similar to the ontology, the schema lists all possible values for such slots. Furthermore, during annotation, the values of these slots in the dialogue state and user or system actions must be selected from a pre-defined candidate list defined in the schema. This helps achieve sanity and consistency in annotations.

We define categorical and non-categorical slots for each domain as shown in Table~\ref{table:schema}. The idea of splitting the slots in MultiWOZ into categorical and non-categorical is not new. Many models have used the number of possible slot values as the classification criterion~\cite{zhang2019find}. Similarly, we classify slots with fewer than 50 different slot values in the training set as categorical, and the others as non-categorical. 

Note that since the Bus and Police domains have very few dialogues in the training set (5 and 145 respectively), the number of possible slot values in this domain does not reflect the true attributes of the slots. Thus, we classify them by referring to similar slots in different domains instead of following the threshold rule.

\subsection{Categorical Slots}

The list of all possible values for categorical slots is built from the corresponding database provided with MultiWOZ 2.1. In addition, we allow ``dontcare" as a special value, which is used when user doesn't have a preference.  We also observe cases where the mentioned value is outside the scope of the database, such as the example below, where MultiWOZ 2.1 specifies ``\$100" as the value for \textit{hotel-pricerange} in the dialogue state. 

\begin{description}[leftmargin=*, labelsep=0.2em, itemsep=0em]
\item[User:] \textit{Well,I want it cheaper than AIRBNB,so how about \$100 a night?}

\item[System:] \textit{Unfortunately, we do not have such specific price ranges, but our options are divided into 3 categories: cheap, moderate or expensive. Which would you prefer?}
\end{description}
Since ``\$100" is not a part of the schema, we use ``unknown" as the slot value in the dialogue state to express that the requirement specified by the user can not be fulfilled by the schema.

\subsection{Non-categorical Slots}

Values of non-categorical slots are extracted from the dialogue history. Due to the typographical errors and slot value paraphrasing, the exact value can not be found in many cases. Some examples are ``el shaddia guest house" being written as ```el shaddai" or ``18:00" being written as ``6pm" in the dialogue utterances.
Since in practice, typographical errors are inevitable and the same value can be mentioned in variable ways, we try to not modify the utterance to keep the dialogue natural. We also allow the presence of more than one value in the dialogue state. During evaluation, a prediction listing either of the listed values is considered correct.

\begin{figure}[t]
\centering
\includegraphics[width=0.42\textwidth]{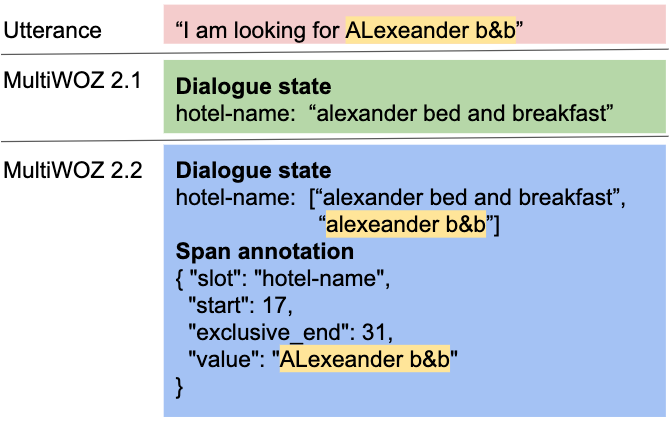}
\setlength{\belowcaptionskip}{-10pt}
\caption{Example of the difference between dialogue state annotation in MultiWOZ 2.1 and MultiWOZ 2.2 and span annotations in MultiWOZ 2.2.}
\label{fig:noncat_slot}
\end{figure}

We use a customized string matching method that takes into consideration the possible typos and alternative expressions to locate all values semantically similar to the annotation. If there are multiple matches, we select the most recently mentioned value and annotate its span. We also add this value to the dialogue state, while preserving the original value. Figure~\ref{fig:noncat_slot} shows the differences between the annotations in MultiWOZ 2.1 and MultiWOZ 2.2. The former only assumes a single value for each slot, even though the slot values can be mentioned in multiple ways and predicting any one of these variants should be considered correct. Thus, in MultiWOZ 2.2, the dialogue state can contain a list of values for a slot: predicting any value in this list is considered correct.

In some cases, slot value is carried over from other slots without being explicitly mentioned in the dialogue. For instance, in the utterance ``I need to book a taxi from the museum to the restaurant", the slot value for \textit{taxi-destination} is copied from the value for \textit{restaurant-name} populated earlier. Instead of annotating the span for \textit{taxi-destination}, we note down the original slot that \textit{taxi-destination} copies its value from. The span annotation for such slots can be obtained by tracing back the copy chain. We posit that this information can be beneficial for state tracking models utilizing a copy mechanism. 

\subsection{User and System Actions}
The user and system action annotations provide a semantic representation of the respective utterances. These annotations were not part of the original MultiWOZ 2.0 release. They were created by \citet{lee2019convlab} and were subsequently added to MultiWOZ 2.1. However, around 5.82\% of turns have missing action annotations. We use crowdsourcing to obtain annotations for these 8,333 dialogue turns (7,339 user and 994 system). The slot names used in dialogue acts are slightly different from the ones used in dialogue state annotations. We rename the slots in the dialogue acts to remove this mismatch.

MultiWOZ 2.1 uses domain-specific prefixes to associate actions with a certain domain. A few dialogue acts also have the ``Booking" prefix, which is used in a few domains including Restaurant, Hotel and Train whenever a reservation is involved. In these cases, it is difficult to identify the domain corresponding to the action since the same prefix is used across many domains. We eliminate the domain and ``Booking" prefixes from the dialogue acts, so that a uniform representation of actions can be used across all domains. To retain the association with the domain, actions for the same domain are grouped together into \textit{frames}, following the representation used by \citet{rastogi2019scalable}.

\subsection{Statistics}
Table~\ref{tab:correction_rate} contains statistics on the corrections in the training, dev, and test sets. We observe that the errors are relatively uniformly distributed across the three splits. Combining all the aforementioned procedures, we modify dialogue states in 17.3\% of the user utterances across 28.2\% of all dialogues. Out of the total modified 12,375 utterance annotations, a majority of the corrections fix the state update inconsistencies described in Section~\ref{sec:inconsis_issues} by listing all the different ways in which a value has been mentioned over the dialogue context in the dialogue state. Of these state updates, 1497, or just over 12\% involved corrections for two or more slots. Missing action annotations were added in a total of 8,333 utterances, whereas pre-existing actions in MultiWOZ 2.1 were verified and fixed for around 10\% of the utterances.

\begin{table}[t]
\centering 
    \begin{tabular}[t]{|C{14mm}|C{20mm}|C{28mm}|}\hline
            \textbf{Dataset} & \textbf{\% of state} & \textbf{\% of dialogues} \\\hline
           train & 17.3 & 27.9 \\ \hline
           dev & 17.3 & 28.7  \\\hline
           test & 17.6 & 29.5 \\\hline
    \end{tabular}
    \setlength{\belowcaptionskip}{-10pt}
    \caption{The ratio of the modified dialogue states (same as the number of user utterances) and  modified dialogues in the training, dev, and test sets.}
    \label{tab:correction_rate}
\end{table}

\section{Additional annotations}\label{sec:additional}
Besides the span annotations, we also add active user intents and requested slots for every user turn. Predicting active user intents and requested slots are two new sub-tasks that can be used to evaluate model performance and facilitate dialogue state tracking. Prediction of active intents or APIs is also essential for efficiency in large-scale dialogue systems which support hundreds of APIs.

\begin{itemize}[leftmargin=*]
    \item \textbf{Active intents:} It specifies all the intents expressed in the user utterance. Note that utterances may have multiple active intents. For example, in ``can i get the college's phone number. i am also looking for a train to birmingham new street and should depart from cambridge looking for a train", the user exhibits the intent both to know more about an attraction and to search for a train.

Based on the action and state annotations, we define a single search intent for the Attraction, Bus, Hotel, and Police domains and a single booking intent for Taxi domain, whereas for the Restaurant, Hotel, and Train domains, both search and booking intents are defined. 

    \item \textbf{Requested slots:} It specifies the slots that the user requests information about from the system. This field is generated based on the user actions in each turn. These annotations find direct applicability in developing  dialogue policy models, since requesting additional information about entities is very common in task-oriented dialogue.
\end{itemize}

\section{Dialogue State Tracking Benchmarks}
\label{sec:benchmark}
Recent data-driven dialogue state tracking models that achieve state-of-the-art performance mainly adopt two classes of methods: span-based and candidate-based. Span-based methods extract values from dialogue history and are suitable for tracking states of non-categorical slots, while candidate-based methods that perform classification on pre-defined candidate lists to extract values are better-suited for categorical slots. 
To test models' performance on both categorical and non-categorical slots, we selected three dialogue state tracking models that use a mixture of both methods to benchmark the performance on the updated dataset: SGD-baseline~\cite{rastogi2019scalable}, TRADE~\cite{WuTradeDST2019}, and DS-DST~\cite{zhang2019find}.

TRADE considers each slot as a mixture of categorical and non-categorical slot. It uses a pointer generator architecture to either generate the slot value from a pre-defined vocabulary or tokens in the dialogue history.
On the contrary, SGD-baseline has separate tracking strategies for categorical and non-categorical slots. It first uses a shared pretrained BERT~\cite{devlin2018bert} to encode a context embedding for each user turn, a slot embedding for each slot, and a slot value embedding for each slot value in the candidate list of the categorical slots. Then, it utilizes linear networks to perform classification for the categorical slot and to find start and end span indices for non-categorical slots. DS-DST is a recently proposed model achieving state-of-the-art performance on MultiWOZ 2.1 using pre-trained BERT. The main difference between DS-DST and SGD-baseline is that the context embedding used in DS-DST is conditioned on the domain-slot information while it is not in SGD-baseline.

We use joint goal accuracy as our metric to evaluate the models' performance. The joint goal accuracy is defined as the average accuracy of predicting all the slot values for a turn correctly. The performance of different models is shown in Table~\ref{tab:result}. In general, we observe similar performance on MultiWOZ 2.1 and MultiWOZ 2.2 across the three models. Table~\ref{tab:result2} compares the joint goal accuracy over only the categorical slots (cat-joint-acc) and only the non-categorical slots (noncat-joint-acc) across all the models. It shows that TRADE and SGD-baseline demonstrate considerably higher performance on non-categorical slots than categorical slots. We infer that it may be caused by the corrections ensuring that the value in the dialogue state is also present in the dialogue history for all non-categorical slots.
\begin{table}[t]
\centering
    \begin{tabular}[t]{|C{15mm}|C{14mm}|C{14mm}|C{14mm}|}\hline
            \textbf{Model} & \textbf{MultiWOZ 2.0} & \textbf{MultiWOZ 2.1} & \textbf{MultiWOZ 2.2}\\\hline
            TRADE & 0.486 & 0.460 & 0.454\\\hline
            SGD-baseline & - & 0.434 & 0.420\\ \hline
            DS-DST & 0.522 & 0.512& 0.517\\ \hline
    \end{tabular}
    \caption{Joint goal accuracy of TRADE, SGD-baseline and DS-DST models on MultiWOZ 2.0, MultiWOZ 2.1 and MultiWOZ 2.2 datasets.}
    \label{tab:result}
\end{table}

\begin{table}[t]
\centering
    \begin{tabular}[t]{|C{23mm}|C{20mm}|C{20mm}|}\hline
            \textbf{Model} & \textbf{Cat-joint-acc} & \textbf{Noncat-joint-acc} \\\hline
            TRADE & 0.628 & 0.666 \\\hline
            SGD-baseline & 0.570 & 0.661 \\ \hline
            DS-DST & 0.706 & 0.701 \\ \hline
    \end{tabular}
    \setlength{\belowcaptionskip}{-10pt}
    \caption{Performance of TRADE, SGD-baseline, and DS-DST models on predicting categorical and non-categorical slots. Cat-joint-acc and noncat-joint-acc denote joint goal accuracy on categorical and non-categorical slots respectively.}
    \label{tab:result2}
\end{table}

\section{Discussion}

The Wizard-of-Oz paradigm is a very powerful technique to collect natural dialogues. However, the process of annotating these dialogues is prone to noise. In this section, we discuss some of the best practices to follow during task-oriented dialogue data collection so as to minimize annotation errors.

It is important to define an ontology or schema before data collection, listing the interface of all the domains and APIs. The schema should identify categorical slots, which have a fixed set of possible values, and the annotation interface should enforce the correctness of these slots. In particular, the interface should only allow the annotator to pick one of the values specified in the schema. For non-categorical slots, the interface should only allow values which have been mentioned in the dialogue history, and display an error otherwise. These simple checks help avoid typographical errors and value paraphrasing issues, discussed in Section~\ref{sec:issues}. The annotation task can be followed by simple validation checks to identify erroneous annotations, which can be fixed by a follow-up crowd-sourcing task. For instance, listing the set of all possible values for every slot in the dataset helped us quickly identify instances listing ``thursday" as the value for a time slot or ``no" as the name of a hotel.

We also observed a few annotations utilizing logical expressions to represent the dialogue state. For instance, some dialogue state annotations utilize string ``cheap$>$moderate" to mean that cheap is preferred over moderate, and ``cinema$|$entertainment$|$museum$|$theatre" to mean that all values separated by ``$|$"are acceptable. However, such values are disproportionately rare in the dataset ($<$1\% of dialogues), thus making it difficult for models to handle such cases. It brings into question how to define a more expressive representation which can support such complex annotations and how we should design the model capable of handling such cases. We hope that as the state tracking technology advances, there will be more focus on this direction. On the other hand, it is important to ensure that such complex constraints are proportionately represented in the dataset if the system is intended to support them.

\section{Conclusion}
MultiWOZ 2.1 \cite{eric2019multiwoz} is an improved version of the MultiWOZ 2.0 dataset, which is extensively used as a benchmark for dialogue state tracking. We identify annotation errors, inconsistencies and ontology related issues which were left unaddressed in MultiWOZ 2.1, and publish a corrected version -- MultiWOZ 2.2. We added a new schema, standardized slot values, corrected annotation errors and standardized span annotations. Furthermore, we annotated active intents and requested slots for each user turn, and added missing user and system actions besides fixing existing ones. We benchmark a few state-of-the-art models on the new dataset: experimental results show that the models' performance is similar between MultiWOZ 2.1 and MultiWOZ 2.2. We hope the cleaned dataset helps make fairer comparisons among models and facilitate research in this field.

\bibliography{acl2020}
\bibliographystyle{acl_natbib}

\end{document}